\documentclass[runningheads]{llncs}
\usepackage[T1]{fontenc}
\usepackage{graphicx}
\usepackage{booktabs}
\usepackage{longtable}
\usepackage{xcolor}
\usepackage{amsmath}
\usepackage{amssymb}
\usepackage{algorithm}
\usepackage{algpseudocode}
\usepackage{url}
\usepackage{mdframed}
\usepackage{subcaption}

\newcommand{\X}{\textbf{X}}
\newcommand{\Z}{\textbf{Z}}

\begin{document}

\title{Neurosymbolic Imitation Learning with Human Guidance: A Privileged Information Approach}

\titlerunning{Neurosymbolic Imitation Learning with Privileged Information}

\author{Nikhilesh Prabhakar*\inst{1} \and
Varun Balaji*\inst{1} \and
Athresh Karanam\inst{1} \and
Kristian Kersting\inst{2} \and
Sriraam Natarajan\inst{1}}

\authorrunning{N. Prabhakar et al.}

\institute{Department of Computer Science, The University of Texas at Dallas, Richardson TX 75080, USA 
\and
Computer Science Department and Centre for Cognitive Science, TU Darmstadt, Germany \email{kristian.kersting@tu-darmstadt.de}}

\maketitle
\begin{abstract}
Imitation learning is widely used for learning to act in complex environments. While pure neural-based methods handle high dimensional data effectively, they suffer from the requirement of large number of samples and are prone to overfitting. Pure symbolic approaches, while generalize well, do not handle high-dimensional data effectively. We propose a neurosymbolic approach that achieves the best of both worlds, i.e, handling high-dimensional data while achieving generalization. The key advantage of our approach is that it can effectively exploit additional privileged information that is available only during training (in our case, gaze data). Our empirical evaluations demonstrate the effectiveness, efficiency and the generalization capability of our proposed approach.

\keywords{Neurosymbolic Learning \and Imitation Learning \and Privileged Information \and }
\end{abstract}
\footnotetext{Preprint accepted at the 6th International Joint Conference on Learning \& Reasoning}

\section{Introduction}
\label{sec:intro}

Skill acquisition by observation has long been addressed in the AI community as Imitation Learning~\cite{janpeters2018algoperspectiveonil}. The key idea is to learn a policy (i.e., a mapping from states to actions) from observations of states and corresponding policy from an expert. This problem has been extensively studied under different conditions -- learning by observation~\cite{SegreDeJong1985}, learning from demonstrations \cite{ArgallChernovaVelosoBrowning2009}, programming by demonstrations~\cite{Calinon2009}, programming by example~\cite{Lieberman2000}, apprenticeship learning~\cite{AbbeelNg2004}, behavioral cloning~\cite{SammutEtAl1992}, learning to act~\cite{Khardon1999}, and some others. 

Recent advances in deep learning has accelerated imitation learning from games by learning from raw image data such as Atari games~\cite{bellemare2013ale,mnih2015dqn}. While effective, these methods learn complex policies that are opaque, complex and uninterpretable. While certain local explanation methods exist, the explanations are typically post-hoc. Symbolic methods~\cite{muggleton1991inductive} on the other hand, learn interpretable, explainable, most importantly generalizable and potentially interactive models. However, these models assume that the input representation is symbolic, an assumption that restrict the adaptation of these methods to large domains. 

Neurosymbolic approaches~\cite{evans2018} bridge this gap between deep learning and symbolic methods and have recently re-attracted significant attention due to the emergence of the test-time scaling of autoregressive models. Inspired by the success of these methods, we develop a neurosymbolic method for imitation learning. The key idea in our approach is to use neural methods to create symbolic inputs from raw image data, and learn interpretable and generalizable probabilistic logic rules to model the policy. 

A key aspect of our method is its ability to use additional information. Advice-taking methods that use additional information from the domain expert have a long and cherished history in AI~\cite{VapnikV09LUPI}. Building on the successes of these methods, we consider additional information during training as privileged information (one that is not available during testing/deployment) to learn probabilistic logic rules. We specifically focus on Atari games, a traditional setting hard for symbolic models, and use gaze data as the privileged information. Our proposed framework, Gaze Guided Relational Approach to Imitation Learning (GRAIL) demonstrates efficacy, effectiveness, and generalization on empirical evaluations against purely neural and purely symbolic methods. Our key observation in this work is that combining ideas from classical AI literature such as neural, symbolic, advice-based and imitation learning can result in a robust, generalizable imitation learning model. We make the following key contributions: (1) We develop a novel neurosymbolic framework that learns interpretable probabilistic logic rules for imitation learning. (2)We consider additional information in the form of human gaze as privileged information and develop a robust learning procedure. (3) Our paper combines ideas from multiple areas of AI -- neural learning, symbolic learning, imitation learning and advice-based learning. (4) We empirically validate that the proposed GRAIL framework is both sample efficient, and learns robust policies while generalizing to an unknown number of objects at test-time. 


The rest of the paper is organized as follows: we next provide the necessary background and related work on the key topics of the paper -- imitation learning, neurosymbolic learning and advice-based learning. We next present our key contribution, the GRAIL framework with necessary examples and notations before presenting our empirical evaluations on three ATARI domains, namely, \textit{Asterix},
\textit{Seaquest} and \textit{Freeway}.

\section{Background and Related Work}


\noindent \textbf{Notation:} Let $o_t \in \mathcal{O} \subseteq \mathbb{R}^{C \times H \times W}$ denote the observation at time step $t$, $a_t \in \mathcal{A}$ represent the corresponding action, and $C$, $H$, and $W$ are the number of (stacked grayscale) channels, the frame height, and the frame width, respectively. We define a trajectory of length $T$ as the sequence $\tau = (o_1, a_1, \dots, o_T, a_T) \in (\mathcal{O} \times \mathcal{A})^T$. We define the policy $\pi_{W} : \mathcal{O} \times \mathcal{A} \to [0, 1]$, parameterized by a vector of weights $W \in \mathbb{R}^M$ and defined over a set of $M$ first-order definite clauses $\{c_i\}_{i=1}^M$, as a function mapping the joint observation space and action space to a probability distribution.

We introduce the rest of the notation as necessary.


\subsection{Markov Decision Processes (MDPs)}

A \textit{Markov Decision Process} (MDP) provides the formal framework for modeling sequential decision-making problems. Formally, an MDP is defined as a tuple $\mathcal{M} = \langle \mathcal{S}, \mathcal{A}, \mathcal{T}, \mathcal{R}, \gamma \rangle$, where $\mathcal{S}$ is a set of states, $\mathcal{A}$ is a set of actions, $\mathcal{T}(s, a, s') = P(s_{t+1} = s' \mid s_t = s, a_t = a)$ is the transition function, $\mathcal{R}(s, a, s')$ is the reward function, and $\gamma \in [0, 1)$ is a discount factor \cite{sutton2018}.

In the reinforcement learning setting, the transition dynamics and reward function are typically unknown, and the agent must learn an optimal policy (and in model-based methods, the transition and reward dynamics) through interaction with the environment. 

\subsection{Imitation Learning (IL)}

Imitation Learning (IL) is a learning paradigm in which an agent acquires a policy by observing expert demonstrations. Given a set of expert trajectories, the goal of IL is to learn a policy $\pi$ that replicates the expert's behavior. This is particularly useful in settings where the reward function is difficult to specify manually, as the expert demonstrations implicitly encode the desired behavior. The two main approaches to IL are \textit{Behavioral Cloning} (BC)~\cite{SammutEtAl1992}, which treats policy learning as a (sequential) supervised learning problem, and \textit{Inverse Reinforcement Learning} (IRL)~\cite{NgRussell2000}, which infers the latent reward function that best explains the observed demonstrations \cite{janpeters2018algoperspectiveonil}. Imitation learning has seen BC-based approaches applied to increasingly high-dimensional tasks. However, such approaches often suffer from covariate-shift. This occurs when small errors lead the agent to states not covered by the training distribution. Interactive frameworks such as DAgger \cite{Ross2011DAgger}, SEARN \cite{Daume2009SEARN} and SMILe \cite{Ross2010SMILe} address this by querying experts during training, they require "human-in-the-loop" simulators and incur high operational costs. Beyond covariate shift, purely neural BC architectures also suffer from two additional structural limitations: (1) they do not generalize to scenes with a varying number of objects, and (2) the learned policy remains opaque, offering no symbolic interpretation of the agent's decision-making. Our work addresses all three limitations by replacing the neural policy with a differentiable symbolic reasoner.

\subsection{Learning using privileged information}
The Learning Using Privileged Information (LUPI) paradigm \cite{VapnikV09LUPI,vapnik15LUPISimilarityControlKT} aims to utilize auxiliary features, $\Z$, - in addition to regular features, $\X$ - that are available during training and not during testing with the goal of improving modeling accuracy, enabling faster model convergence, and improving robustness to noise, among others. LUPI methods can be broadly categorized into two complementary categories: distillation-based transfer \cite{hinton2015distillation} and PI-inference-based.

Distillation-based LUPI methods use teacher-student pairs to train a teacher model with access to PI and transfer its information to a student that does not have access to PI. This line of work hinges upon the idea that soft targets from such a teacher can mimic the benefits of PI at test time, as shown analytically by \cite{lopez2015unifying}. In effect, this provides a model-agnostic framework to incorporate PI, where the teacher captures the PI which then implicitly guides the student during training through the distillation process \cite{Hoffman2016LUPIHallucination}. This generic formulation is adapted by several LUPI algorithms \cite{markov2016LUPISpeech,GarciaMM20LUPIAdversarial,YangSRBV22LUPILearningToRank,Yan2023LUPIGBT}.

PI-inference-based LUPI methods, on the other hand, treat $\Z$ as latent variables to be inferred at test-time. These methods are particularly effective when the PI exerts complex and instance-specific influence on the target variable, allowing fine-grained calibrated predictions and explicit probabilistic semantics. In particular, they are effective at dealing with aleatoric uncertainty, especially in the presence of heteroscedastic noise \cite{collier22aTRAM,Lambert2018LUPIHeteroscedastic}, making them ideal candidates for dealing with heteroscedasticity inherent to human demonstrations \cite{Caldarelli2023HeteroscedasticGPLFD}.

Our work belongs to the latter category. However, instead of training a predictive model on the joint input space of $(\X, \Z)$ and marginalizing $\Z$ out at test-time, we propose using the PI to augment the input space directly.  

\subsection{Treating gaze as privileged information}
Treating gaze as an auxiliary signal can allow the agent to learn a model that focuses on causally relevant features without requiring additional expert interaction. Although not explicitly categorized under the LUPI paradigm, much prior work has used gaze as a privileged attention signal available at train time. Gaze-based regularization methods penalize the policy for attending to regions inconsistent with human gaze, either by modifying the loss \cite{Saran2020CGL,Banayeeanzade2025GABRIL} or by modulating dropout within convolutional layers \cite{Chen2019GMD}. A second family of methods directly filters or crops the input observations using a predicted gaze saliency map, instantiated variously as masking \cite{Zhang2018AGIL,liang2024visarl}, observation cropping \cite{Kim2020}, and gaze concatenation in drone racing and autonomous driving contexts \cite{Pfeiffer2022}. A third family adopts a multi-objective architecture, jointly predicting gaze and control commands so that gaze acts as an inductive bias during training \cite{Thakur2023GRIL}. Critically, all of these methods apply gaze guidance to a neural policy, preserving the opacity inherent to such architectures. In contrast, we apply gaze as an attention prior over the symbolic grounding stage, guiding which relational atoms the perception module attends to, rather than regularizing pixel-level feature maps.

\subsection{Relational and Neurosymbolic Imitation Learning}

Relational representations offer a natural solution to the generalization problem, as policies defined over object relations rather than fixed-size feature vectors can operate on a varying number of entities \cite{dzeroski2001rrl}. Inductive Logic Programming (ILP) provides a framework for learning relational policies by inducing first-order rules from demonstrations. Formally, given background knowledge $B$ and a language bias $\mathcal{L}$, ILP seeks a hypothesis $H \subseteq \mathcal{L}$ of definite clauses of the form $\alpha 
\leftarrow \alpha_1, \ldots, \alpha_m$, where each atom is an $n$-ary predicate over variables or constants~\cite{lloyd1987}. While expressive and interpretable, classical ILP requires clean, pre-specified symbolic inputs and cannot operate 
directly on high-dimensional visual observations. differentiable ILP ($\partial$ILP) addresses this by replacing hard symbolic inference with gradient-based optimization, enabling rule learning via backpropagation~\cite{evans2018}. More recent neurosymbolic approaches couple a neural perception front-end with a differentiable symbolic reasoner that enables training from raw observations 
The Neural Symbolic Forward Reasoner (NSFR) \cite{shindo2021neuro} 
performs differentiable forward-chaining inference over a library of first-order logic (FOL) clauses, using soft t-norm and t-conorm operators to approximate logical conjunction and disjunction in a fully differentiable manner. While NSFR has been applied in supervised classification settings, its application to imitation learning from visual demonstrations with gaze-guided symbolic grounding has not been explored. Our work bridges this gap by training an NSFR-based policy on the Atari-HEAD dataset, using human gaze to supervise the neural grounding module that maps raw frames to probabilistic relational atoms.
\section{Neurosymbolic Imitation Learning using Privileged Information}
\label{sec:main}

\begin{figure}[t]
    \centering
    \includegraphics[width=\textwidth]{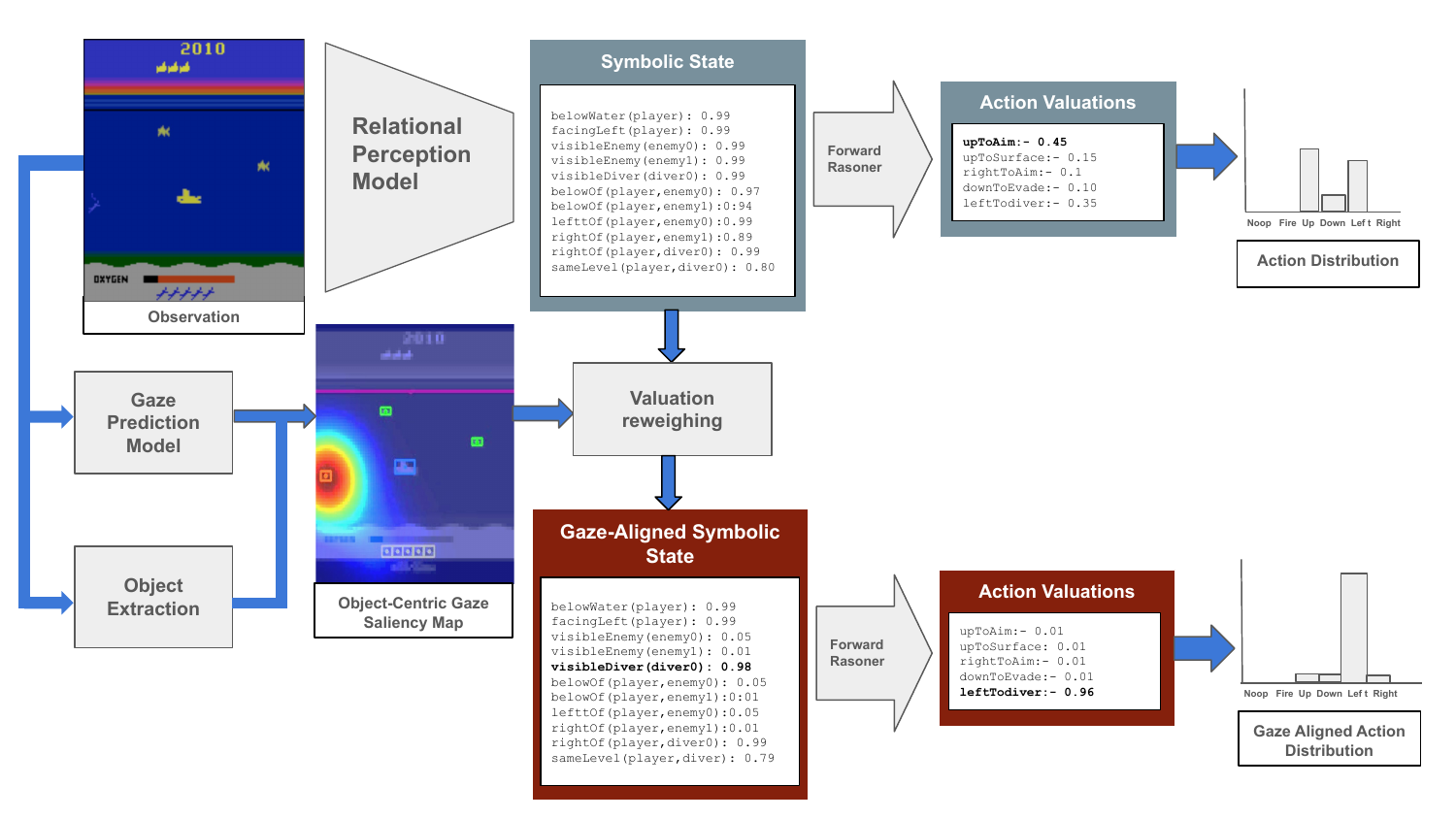}
    \caption{\textbf{Neurosymbolic Imitation Learning (NESY-IL) Architecture.} The framework integrates high-dimensional visual perception with structured symbolic reasoning. A perception model extracts a set of grounded atoms from raw observations. Concurrently, a gaze prediction model and an object extractor provide object-level saliency, which is used to reweigh the states and align them with human attention.  This gaze-aligned symbolic representation is then processed by a differentiable forward reasoner to generate action valuations and a final action distribution that reflects expert attentional biases.}
    \label{fig:nesy-il-architecture}
\end{figure}
\vspace{-1em}
We formally define the learning problem as follows 

\vspace{-0.5em}
\begin{mdframed}[innertopmargin=5pt, innerbottommargin=5pt, skipabove=4pt, skipbelow=4pt]
\noindent\textbf{Given:} A dataset $\mathcal{D} = \{\tau_i\}_{i=1}^{N}$ of expert 
trajectories, its corresponding privileged gaze heatmap 
$G_t \in \Delta^{H \times W}$~\cite{VapnikV09LUPI}, and a rule base 
$\mathcal{B} = \{c_j\}_{j=1}^{M}$ of first-order definite clauses encoding expert 
domain knowledge over grounded atoms $\mathcal{P}$.

\smallskip
\noindent\textbf{To Do:} Learn a policy $\pi_W$ such that the rollout regret $J(\pi^*)- J(\pi_W)$ to the expert policy $\pi^*$ is minimal.
\end{mdframed}

To this end, we propose Gaze-guided Relational Imitation Learning (GRAIL), 
which integrates three tightly coupled modules: a neural perception module that grounds high-dimensional visual observations into (probabilistic) atoms; a gaze-conditioned attention mechanism that modulates atom valuations using privileged human eye-tracking data; and a differentiable forward-chaining reasoner that learns an interpretable, rule-weighted policy via behavior cloning. Figure~\ref{fig:nesy-il-architecture} illustrates the full pipeline. As far as we are aware, this is the first neurosymbolic learning framework that can exploit (propositional) additional knowledge, in this case gaze information, to learn symbolic policies. 
 
\subsection{Neurosymbolic Grounding via Probabilistic Atom Valuation}

Note that unlike other typical symbolic learning methods, we do not assume an input representation that is also symbolic. Instead, we assume that the input is a set of raw images. Consequently, the first key challenge in our imitation learning is \textit{grounding} of symbols: mapping raw pixel observations to a structured representation suitable for symbolic inference. The grounded state of a domain is defined by a finite set of grounded atoms $\mathcal{P} = \{p_1, \ldots, p_{|\mathcal{P}|}\}$, where each atom $p_i$ is an instantiated predicate over domain entities (e.g., $\texttt{close}(\texttt{player}, \texttt{enemy})$).


We train a neural perception network $f_\theta : \mathbb{R}^{C \times H \times W} \to [0, 1]^{|\mathcal{P}|}$ to produce a probabilistic valuation vector:
\begin{equation}
    \mathbf{v}^{(0)}_t = f_\theta(o_t) \in [0, 1]^{|\mathcal{P}|}
\end{equation}
where $v^{(0)}_{t,i}$ denotes the inferred truth probability of atom $p_i$ at timestep $t$. Inspired by well-known probabilistic logic frameworks~\cite{Poole93,Sato94,Kimmig07}, probabilities are assigned to ground atoms and first-order clauses are learned as unweighted clauses.

To supervise this module, we first apply a non-differentiable object extraction oracle (OCAtari) 
to parse expert trajectories into ground-truth binary valuations $\mathbf{y}_t \in \{0, 1\}^{|\mathcal{P}|}$. The perception network is then trained independently by minimizing the negative log-likelihood (NLL) loss:
\begin{equation}
    \mathcal{L}_{\text{NLL}}(\theta) = -\frac{1}{|\mathcal{P}|} \sum_{i=1}^{|\mathcal{P}|} y_{t,i} \log v^{(0)}_{t,i}
\end{equation}
Once trained, $f_\theta$ serves as a fixed symbolic feature extractor, decoupling visual perception from downstream logical reasoning.

\vspace{-0.5em}
\begin{algorithm}[t]
\caption{GRAIL: Gaze-guided Relational Imitation Learning}
\label{alg:grail}
\begin{algorithmic}[1]

\Require Expert trajectories $\mathcal{D} = \{\tau_i\}_{i=1}^{N}$, gaze heatmaps $\{G_t\}$, atom set $\mathcal{P}$, rule base $\mathcal{B}$, max steps $T_{\max}$

\Function{GRAIL}{$\mathcal{D}, \{G_t\}, \mathcal{P}, \mathcal{B}$}
    \For{each $(o_t, \mathbf{y}_t)$ from $\mathcal{D}$} \Comment{Train perception module with ground-truth atom labels}
        \State $\mathcal{L}_{\text{NLL}} \leftarrow -\sum_i y_{t,i} \log v^{(0)}_{t,i}$
        \State Update $\theta$ via $\nabla_\theta \mathcal{L}_{\text{NLL}}$
    \EndFor
    \For{each $(o_t, G_t)$ from $\mathcal{D}$} \Comment{Train gaze prediction module}
        \State $\mathcal{L}_{\text{gaze}} \leftarrow D_{\text{KL}}\!\left(G_t \;\Vert\; g_\phi(o_t)\right)$
        \State Update $\phi$ via $\nabla_\phi \mathcal{L}_{\text{gaze}}$
    \EndFor
    \State Freeze $\theta$, $\phi$; initialize $W$ \Comment{Begin policy learning}
    \For{each $(o_t, a_t)$ from $\mathcal{D}$}
        \State $\mathbf{v}^{(g)}_t \leftarrow$ \Call{GazeModulate}{$o_t, \mathcal{P}$} \Comment{Gaze-filtered grounding}
        \State $\mathbf{v}^{(T_{\max})} \leftarrow$ \Call{ForwardChain}{$\mathbf{v}^{(g)}_t, \mathcal{B}, W$} \Comment{Differentiable inference}
        \State $s_c \leftarrow \max_{r \in \mathcal{R}_c} v^{(T_{\max})}(r) \quad \forall c \in \mathcal{A}$
        \State $\mathcal{L}_{\text{policy}} \leftarrow -\log \pi_W(a_t \mid o_t)$
        \State Update $W$ via $\nabla_W \mathcal{L}_{\text{policy}}$
    \EndFor
    \State \Return $W$
\EndFunction    



\end{algorithmic}
\end{algorithm}
\vspace{-0.5em}

\subsection{Privileged Gaze Modulation of Atom Valuations}

Even after grounding, the valuation vector $\mathbf{v}^{(0)}_t$ may assign non-trivial truth probabilities to atoms involving task-irrelevant background entities, introducing spurious correlations into downstream rule learning. We resolve this credit assignment problem via \textit{privileged information} in the form of human eye-tracking data, available only during training under the LUPI paradigm \cite{VapnikV09LUPI}.

\paragraph{Gaze Predictor.} Prior to policy training, we train a dedicated gaze prediction network $g_\phi : \mathbb{R}^{1 \times H \times W} \to \mathbb{R}^{H \times W}$ that maps a single grayscale frame $o_t \in \mathbb{R}^{1 \times H \times W}$ to a spatial probability distribution over fixation locations. Ground-truth gaze coordinates are converted into 2D Gaussian heatmaps $G_t$, and $g_\phi$ is optimized by minimizing the Kullback-Leibler divergence:
\begin{equation}
    \mathcal{L}_{\text{gaze}}(\phi) = D_{\text{KL}}\!\left(G_t \;\Vert\; g_\phi(o_t)\right)
\end{equation}

\paragraph{Atom-Level Gaze Scoring.} During policy training, $g_\phi$ is frozen and produces a normalized gaze heatmap $\hat{G}_t$, where $\sum_{x,y} \hat{G}_t(x, y) = 1$. An atom $p_i$ in $\mathcal{P}$ may reference one or more entities $e_i^{(1)} \cdots e_i^{(k)}$, each with an associated bounding box $\beta_i$. We compute a per-entity saliency score by calculating a gaze mass as follows
\begin{equation}
    s_i = \sum_{(x,y) \in \beta_i} \hat{G}_t(x, y) \in [0, 1]
\end{equation}
For atoms referencing a single entity, the scalar gaze score is $s_i = s_i^{(1)}$. For atoms referencing multiple entities, we aggregate using the product t-conorm
\begin{equation}
    s_i = 1 - \prod_{j=1}^n (1-s_i^{(j)})
\end{equation}

\paragraph{Gaze-Modulated Valuation.} The gaze score is applied as a soft multiplicative scaling over the initial valuation vector, yielding the gaze-modulated symbolic state:
\begin{equation}
    v^{(g)}_{t,i} = v^{(0)}_{t,i} \cdot s_i
\end{equation}

The intuition is that we essentially reweigh the probabilities of the ground atoms based on the gaze information. If the gaze heatmap covers the grounding object, the corresponding probability is increased else it is decreased. This is akin to the idea of relevance information used in earlier ILP learning systems~\cite{Shavlik2009Onion} where the importance of a grounded predicate can be decreased or increased using domain knowledge thus changing the score the clause induced by the ground atom. We instead change the probabilities of these atoms.

\subsection{Differentiable Forward-Chaining Inference for Policy Learning}

The gaze-filtered valuation vector $\mathbf{v}^{(g)}_t$ is passed to a differentiable logic reasoning engine, implemented via the Neurosymbolic Forward Reasoner (NSFR). 
NSFR performs forward-chaining inference over a predefined rule base $\mathcal{B}$, a set of first-order definite clauses of the form:
\begin{equation}
    \alpha_0 \leftarrow \alpha_1, \alpha_2, \ldots, \alpha_m
\end{equation}
where each $\alpha_j$ is a grounded FOL atom. Discrete Boolean inference is relaxed into continuous arithmetic via fuzzy logic semantics: logical conjunction ($\wedge$) is approximated by the product T-norm, and disjunction ($\vee$) by the corresponding T-conorm, enabling end-to-end differentiation through the inference graph.

\paragraph{Forward Chaining.} Starting from the initial facts $\mathbf{v}^{(g)}_t$, the engine performs $T_{\max}$ deductive steps, iteratively updating atom valuations by applying the weighted rule base. This yields a final valuation tensor $\mathbf{v}^{(T_{\max})}_t$ encoding the inferred truth probabilities of all action predicates.

\paragraph{Behavior Cloning Objective.} Let $\mathcal{R}_c \subseteq \mathcal{B}$ denote the subset of clauses with action $c$ as their head. The aggregated action score is:
\begin{equation}
    s_c = \max_{r \in \mathcal{R}_c} v^{(T_{\max})}_{t}(r)
\end{equation}
producing an unnormalized score vector $\mathbf{s} \in [0,1]^{|\mathcal{A}|}$. The learnable rule weight matrix $W$ is optimized by minimizing the negative log-likelihood of the expert action under the induced policy $\pi_W(a \mid o_t)$:
\begin{equation}
    \mathcal{L}_{\text{policy}}(W) = -\sum_{t=1}^{T} \log \pi_W(a_t \mid o_t)
\end{equation}
Because $\mathbf{v}^{(g)}_t$ has been filtered by the privileged gaze mechanism, backpropagation updates $W$ exclusively on task-critical relational features, preventing the formation of spurious correlations with background distractors and improving both sample efficiency and out-of-distribution generalization.

\subsection{GRAIL Framework}

Putting all of these together, we get the GRAIL framework that is outlined in Figure~\ref{alg:grail}. To reiterate, the framework obtains raw images as inputs, converts them to probabilistic grounded atoms using the perception module. Then given the gaze information for the corresponding raw image, the probabilities of the ground atoms are reweighted accordingly. Next, these grounded atoms are used to learn relational policies using a forward reasoner that employs a behavioral cloning objective. The final policies obtained in this fashion are inrepretable, explainable and most importantly, as we demonstrate, generalizable. As far as we are aware, this is the first framework to implement privileged information based re-weighting to learn a neurosymbolic policy. We demonstrate that the policies learned using this framework are effective, efficient and generalizable.

\section{Experimental Evaluation}

Our experiments explicitly aim at answering the following questions:
\begin{itemize}
    \item \textbf{Q1 --- Effectiveness:} Does GRAIL learn a competitive game-playing policy compared to the baseline methods while effectively leveraging gaze data?
    \item \textbf{Q2 --- Sample Efficiency:} Does access to human gaze data during training improve performance in low-data regimes?
    \item \textbf{Q3 --- Generalization:} Does the symbolic state representation generalize better than pixel-based approaches when the test environment contains unseen objects?
\end{itemize}
Our code is publicly available at \url{https://anonymous.4open.science/status/Gaze-Based-Neurosymbolic-Imitation-Learner-665E}.

\subsection{Domain}

We evaluate \textsc{GRAIL} on three Atari 2600 environments, Freeway, Seaquest, and Asterix, drawn from the Atari Human Eye-Tracking and Demonstration (Atari-HEAD) dataset~\cite{zhang2019atarihead}. Atari-HEAD creates gameplay recordings with high-precision eye-tracking data and human action demonstrations across 20 games and multiple human subjects, comprising over 100 hours of recorded play. To ensure consistency, we 
restrict our experiments to the trajectories of a single human player and use only primitive actions, excluding key combinations for a controlled demonstration dataset.

\begin{table}[t]
\caption{Mean Game Score across Games}
\centering
\resizebox{\textwidth}{!}{
\begin{tabular}{lccccc}
\toprule
\textbf{Game} & \textbf{BC} & \textbf{AGIL} & \textbf{BC+Mask} & \textbf{NSFR-IL} & \textbf{GRAIL} \\
\midrule
Asterix  & $306.00 \pm 223.75$ & $228.00 \pm 175.26$ & $251.00 \pm 153.46$ & $2519.00 \pm 1488.65$ & $\mathbf{6308.00 \pm 3765.63}$ \\
Seaquest & $139.20 \pm 39.18$  & $192.80 \pm 60.89$  & $182.00 \pm 60.82$  & $454.20 \pm 272.71$   & $\mathbf{501.60 \pm 216.84}$ \\
Freeway  & $29.74 \pm 1.20$    & $29.02 \pm 1.05$    & $27.52 \pm 1.54$                & $24.58 \pm 1.28$                    & $\mathbf{30.02 \pm 0.41}$ \\
\bottomrule
\end{tabular}}
\label{tab:results}
\end{table}

\subsection{Baselines}
We compare our algorithm, \textbf{GRAIL}, against four baselines --- (i) a purely neural behavioral cloning baseline with no gaze information (\textbf{BC}); (ii) a gaze-augmented 
neural baseline that masks pixel observations with a predicted gaze saliency map, requiring the gaze predictor at test time (\textbf{AGIL}); (iii) a 
gaze-augmented neural baseline that multiplies input observations by a 
gaze-derived attention mask to suppress task-irrelevant pixel regions 
(\textbf{BC+Mask}); and (iv) an ablation of our own framework that removes 
the gaze modulation module entirely, substituting unmodulated atom valuations 
$\mathbf{v}^{(0)}_t$ for the gaze-filtered $\mathbf{v}^{(g)}_t$ 
(\textbf{NSFR-IL}), directly isolating the contribution of privileged gaze 
information within the neurosymbolic setting.

\subsection{Experimental Results}
\subsubsection{Q1. Effectiveness}
To assess whether the neurosymbolic architecture learns a competitive 
game-playing policy, we compare mean game scores across all methods on the three domains, reported across 50 evaluation seeds in 
Table~\ref{tab:results}. GRAIL achieves the highest mean score in every domain. On Asterix and Seaquest, both neurosymbolic methods considerably outperform the neural baselines. Freeway is near-saturated, so all methods cluster tightly; here NSFR-IL falls below the neural baselines, and gaze modulation is what keeps GRAIL marginally ahead. Comparing GRAIL directly to NSFR-IL demonstrates the significant marginal gain of gaze modulation in our framework, as the two methods share identical architectures and rule bases, differing only in whether gaze modulation is applied. This confirms the effectiveness of GRAIL's neurosymbolic architecture as well as its ability to effectively leverage gaze data.


\subsubsection{Q2. Sample Efficiency}
\begin{figure}[t]
\centering
\includegraphics[width=\linewidth]{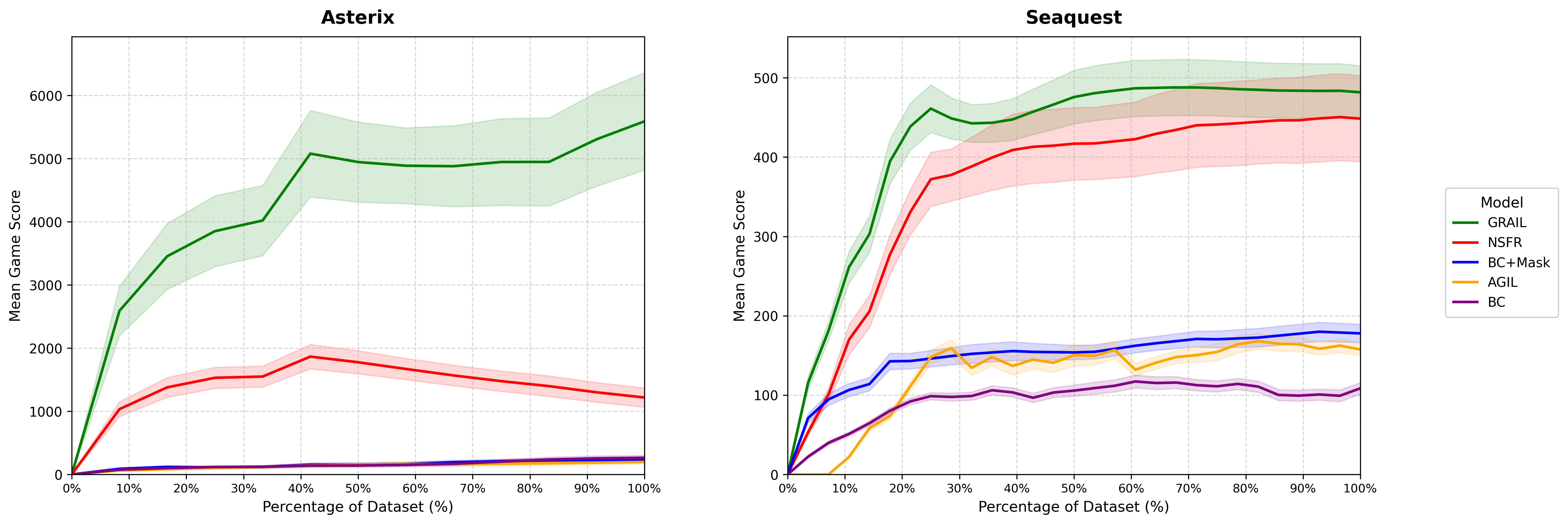}
\caption{\textbf{Comparison of sample efficiency in Atari games.} Mean Game Score is plotted against the percentage of training dataset used for two domains, Asterix (left) and Seaquest (right). In both the domains, GRAIL achieves higher performance earlier and converges to a higher asymptote.}
\label{fig:all_efficiency}
\end{figure}

To assess whether access to human gaze data improves performance in 
low-data regimes, we compare the learning curves of all methods as a 
function of the number of training demonstrations, shown in 
Figure~\ref{fig:all_efficiency}. The learning curves show that GRAIL 
converges to a higher asymptote than all baselines in Asterix and Seaquest, and 
does so with substantially fewer samples. Most strikingly, GRAIL 
trained on $10\%$ of the available data already outperforms NSFR-IL 
trained on the full dataset. Neural baselines plateau at lower 
performance levels and do not recover with additional data, suggesting 
that the bottleneck is representational rather than statistical.
A notable phenomenon is the performance degradation of NSFR-IL on 
Asterix as the number of samples grows. We believe that this reflects a gap between the expert rules crafted and the behavior of the human gameplay data. As more samples are observed, the model increasingly fits spurious correlations in the training data that do not generalize to evaluation. GRAIL does not 
exhibit this degradation, suggesting that gaze modulation acts as a regularizer. By down-weighting atom valuations for entities the human expert does not attend to, it 
prevents the rule weights from latching onto relational features that co-occur with correct actions in the training data but are not relevant to the policy. In effect, human attention provides a supervisory signal that steers the learner away from spurious correlations that additional data would otherwise support.

\begin{figure}[t]
    \centering
    \includegraphics[width=0.85\linewidth]{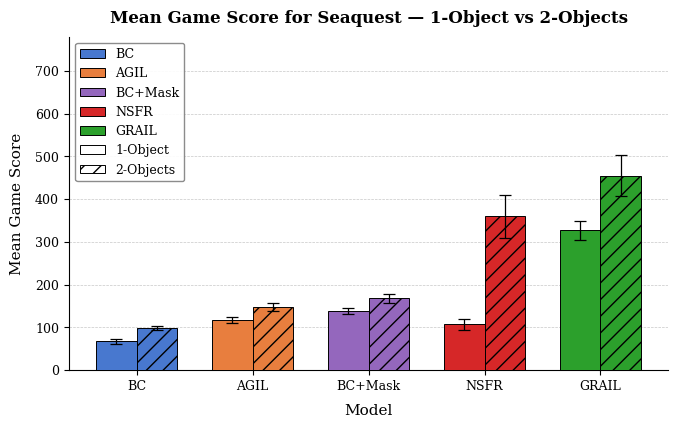}
    \caption{\textbf{Generalization across variable object counts in Seaquest.} Mean game scores are reported for configurations trained on a dataset containing a maximum of one and two objects per object type.}
    \label{fig:object_generalization}
\end{figure}
\vspace{-1em}

 \subsubsection{Q3. Generalization}

To assess whether the symbolic state representation generalizes better than pixel-based approaches when the test environment contains unseen objects, we train all methods on datasets containing a maximum of one and two objects per object type in Seaquest. The results are presented in Figure~\ref{fig:object_generalization}.
Neural baselines --- BC, AGIL, and BC+Mask --- show only marginal improvement when moving from the 1-object to the 2-objects training regime, with mean game scores remaining below 200 across both conditions. Both neurosymbolic methods, on the other hand, benefit considerably from the richer training distribution. NSFR-IL improves substantially in the 2-objects setting, and GRAIL improves further still, achieving the highest mean reward overall. This is consistent with the compositionality of first-order relational representations. The policy is defined over object-level predicates rather than fixed-size feature vectors, so it naturally extends to scenes containing a greater number of entities. The additional advantage of GRAIL over NSFR-IL in the 2-objects setting further suggests that gaze modulation helps the reasoner maintain focus on task-critical atoms as scene complexity increases.

In summary, firstly, our experimental results across three domains show that our framework, GRAIL, outperforms both the neurosymbolic as well as the purely neural baseline methods. Secondly, on Asterix and Seaquest, NSFR-IL surpasses the neural baselines in effectiveness (Q1), sample efficiency (Q2) and generalization (Q3), indicating that the neurosymbolic methods help regardless of PI. Finally, our results show that the proposed gaze modulation effectively and consistently leverages PI to achieve higher overall performance (Q1), faster convergence in low-data regimes (Q2), and generalizes to unknown number of objects (Q3) at test-time through a combination of the neurosymbolic architecture and PI.  


\section{Conclusion}
We presented GRAIL, a gaze-guided neurosymbolic imitation learning framework that is capable of learning from raw image data and additional privileged information in the form of gaze maps. This is the first-of-its-kind framework that effectively employs privileged information that is available during training but not during deployment. Note that because we considered raw game playing data, gaze information served as a natural privileged information. Our framework is certainly not limited to the use of only gaze information. Additional privileged information, if available can be easily integrated to the framework as they are mainly used to reweigh the probabilities.
Another key aspect of the system is that it is a plug and play framework and one can employ any differentiable learner for learning the policies, any perception module capable of extracting probabilistic facts and as mentioned earlier, any form of privileged information for reweighing the probabilities.
Our learning system is capable of learning effective, efficient, explainable and generalizable policies as we demonstrated empirically in multiple image-based domains, traditionally hard ones for symbolic learning.

Our predicate vocabulary and rules are currently hand-crafted per domain; learning them via predicate-invention methods ~\cite{silver2023predicates}
is an immediate next step. Scaling the algorithms to larger problems including clinical decision-support is also an interesting future directions. Integrating other forms of domain knowledge in the form of qualitative constraints, preferences and shaping functions is another potential direction. Similarly, incorporating gaze information as other forms including regularization as well as implementing rule learners are immediate next steps. Allowing for richer human interaction requires integration with a LLM-based interaction system, another interesting direction. Finally, developing imitation learning systems that knows what it knows and solicits information (whether it is gaze or preferences or shaping functions etc) when it does not know,  is one of the most exciting questions in building true human-allied learning systems that learn to act from observations.

\bibliographystyle{splncs04}
\bibliography{references}

\appendix

\label{apd:first}

\end{document}